\documentclass[conference]{IEEEtran}
\IEEEoverridecommandlockouts
\usepackage{cite}
\usepackage{amsmath,amssymb,amsfonts}
\usepackage{algorithmic}
\usepackage{graphicx}
\usepackage{textcomp}
\usepackage{xcolor}

\usepackage{algorithmic}
\usepackage{algorithm}
\usepackage{comment}
\usepackage{ifthen}
\usepackage{fancyhdr}
\def\BibTeX{{\rm B\kern-.05em{\sc i\kern-.025em b}\kern-.08em
    T\kern-.1667em\lower.7ex\hbox{E}\kern-.125emX}}
\begin{document}
\fancyhead[CO,CE]{IEEE Copyright Notice Copyright (c) 2023 IEEE Personal
use of this material is permitted. 
Accepted to be published in 2022 IEEE International Conference on Knowledge Graph (ICKG).
DOI: 10.1109/ICKG55886.2022.00052}

\title{QUBO Decision Tree: Annealing Machine Extends Decision Tree Splitting\\
}

\makeatletter
\newcommand{\linebreakand}{%
  \end{@IEEEauthorhalign}
  \hfill\mbox{}\par
  \mbox{}\hfill\begin{@IEEEauthorhalign}
}
\makeatother

\ifthenelse{1>0}{
\author{
\IEEEauthorblockN{Koichiro Yawata}
\IEEEauthorblockA{\textit{Research and Development Group} 
\textit{Hitachi Ltd.}\\
Kokubunji-shi, Japan \\
koichiro.yawata.rt@hitachi.com}
\and
\IEEEauthorblockN{Yoshihiro Osakabe}
\IEEEauthorblockA{\textit{Research and Development Group} 
\textit{Hitachi Ltd.}\\
Kokubunji-shi, Japan \\
yoshihiro.osakabe.fj@hitachi.com}

\linebreakand 
\IEEEauthorblockN{Takuya Okuyama}
\IEEEauthorblockA{\textit{Research and Development Group} 
\textit{Hitachi Ltd.}\\
Kokubunji-shi, Japan \\
takuya.okuyama.mn@hitachi.com}
\and
\IEEEauthorblockN{Akinori Asahara}
\IEEEauthorblockA{\textit{Research and Development Group} 
\textit{Hitachi Ltd.}\\
Kokubunji-shi, Japan \\
akinori.asahara.bq@hitachi.com}
}
}
{
  \author{
    \IEEEauthorblockN{1\textsuperscript{st} Anonymous}
    \IEEEauthorblockA{\textit{Anonymous: dept. name of organization} 
    \textit{.}\\
    City, Country \\
    email address or ORCID}
    \and
    \IEEEauthorblockN{2\textsuperscript{nd} Anonymous}
    \IEEEauthorblockA{\textit{Anonymous: dept. name of organization} 
    \textit{.}\\
    City, Country \\
    email address or ORCID}
    \linebreakand 
    \IEEEauthorblockN{3\textsuperscript{rd} Anonymous}
    \IEEEauthorblockA{\textit{Anonymous: dept. name of organization} 
    \textit{.}\\
    City, Country \\
    email address or ORCID}
    \and
    \IEEEauthorblockN{4\textsuperscript{th} Anonymous}
    \IEEEauthorblockA{\textit{Anonymous: dept. name of organization} 
    \textit{.}\\
    City, Country \\
    email address or ORCID}
  }
}

\maketitle

\begin{abstract}
This paper proposes an extension of regression trees by quadratic unconstrained binary optimization (QUBO). 
  Regression trees are very popular prediction models that are trainable with tabular datasets,
  but their accuracy is insufficient because the decision rules are too simple.
  The proposed method extends the decision rules in decision trees to multi-dimensional boundaries.
  Such an extension is generally unimplementable because of computational limitations,
  however, the proposed method transforms the training process to QUBO,
  which enables an annealing machine to solve this problem.
\end{abstract}

\begin{IEEEkeywords}
annealing machine, quadratic unconstrained binary optimization (QUBO),
decision tree, regression tree, MCMC
\end{IEEEkeywords}

\begin{figure*}
  \centering
  \includegraphics[width=1.0\linewidth,page=3]{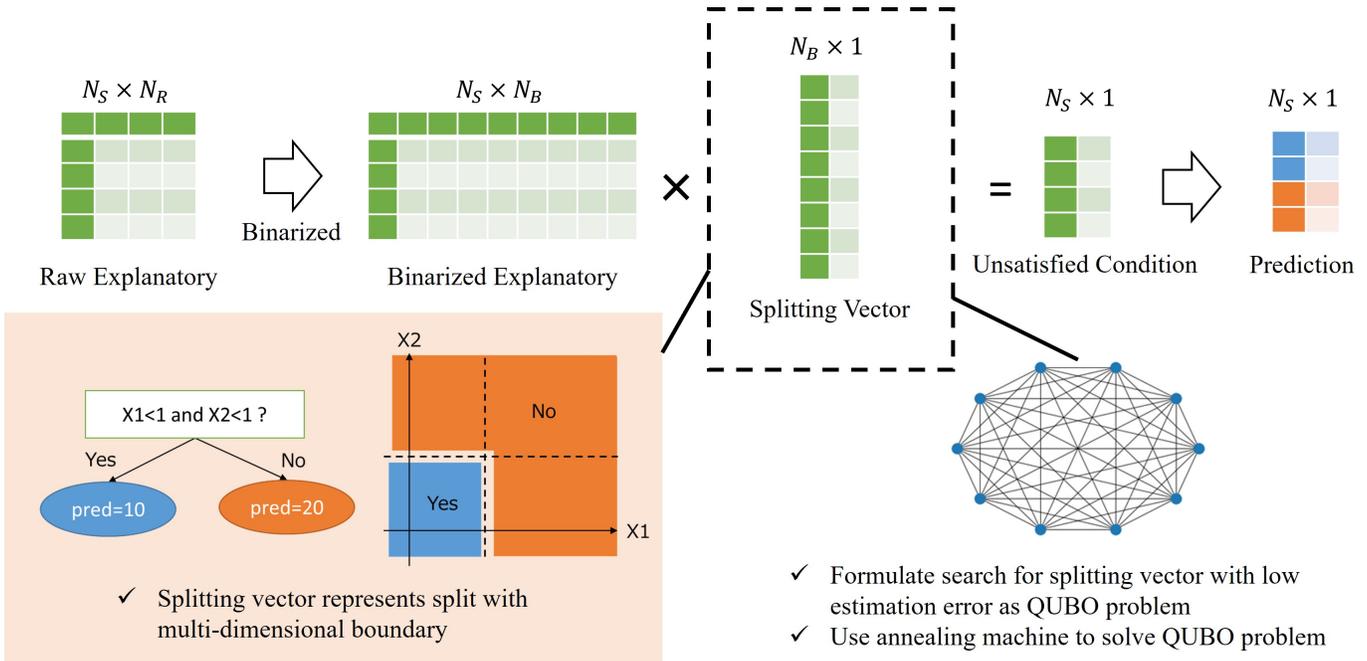}
  \caption{Overview.
  We formulate the search problem of finding splitting conditions
  in regression trees as a QUBO problem
  that can be solved by an annealing machine.
  In addition, we extend regression trees to search more complicated conditions,
  which are defined by multi-dimensional boundaries.
  }\label{fig:test7}
\end{figure*}

\section{Introduction}
Annealing machines have an advantage in solving combinatorial optimization problems
because they can handle various possible states simultaneously \cite{martovnak2004quantum}.
In recent years, the performance and convenience of annealing machines have been improved \cite{shin2014quantum}, \cite{ibm2021}.
Quantum annealing has attracted significant attention as one of the quantum computers \cite{kadowaki1998quantum}, \cite{Justin1999}.
As a result, in addition to applications for typical combinatorial optimization problems
such as the travelling salesman problem and the maximum cut problem, 
it is expected that other algorithms using annealing machines will be developed \cite{prasanna2021qubo}.

Regression trees are decision trees for a regression task,
where decision trees are a machine learning algorithm that estimates
an objective variable from an explanatory variable \cite{safavian1991survey}.
Decision-tree-based methods such as XGBoost \cite{chen2016xgboost} and
LightGBM \cite{ke2017lightgbm} are widely used
because of their high performance \cite{Bennett2007}, \cite{He2014} and accountability \cite{breiman2001random}.
In decision trees, sample split using various feature values are created iteratively
to reduce the prediction error using greedy algorithms.

However, because of computational limitations,
regression trees allow the use of only one feature for each splitting,
and that do not support splitting with complex conditions using multiple features
\cite{loh2011classification,grabczewski2005feature}.
The reason is that the size of the search space increases exponentially with the number of features used,
which makes the optimization more difficult.
This limitation may lead to non-effective splitting,
which has a negative impact on a model's accuracy and interpretation.

To overcome this problem, we consider splitting optimization techniques for regression trees
by using simulated annealing \cite{kirkpatrick1983optimization},
which is a type of annealing machines.
The annealing technique is a stochastic approximation algorithm for finding a state
such that a given function is minimized.
In simulated annealing,
convergence is achieved by making stochastic transitions based on a Markov chain Monte Carlo (MCMC) algorithm.

To optimize regression trees with simulated annealing, we propose two techniques.
First, for the search problem of finding a splitting to reduce the prediction error
in regression trees, we formulate the problem as a quadratic unconstrained binary optimization (QUBO) problem
that can be solved by an annealing machine.
Second, we extend regression trees to search more complicated conditions.
In conventional methods,
only one feature value can be used for splitting samples.
In contrast, the proposed method enables decision rules with multiple features
as shown in Fig. \ref{fig:test7}.
Splitting using multiple features is expected to reduce the number of splitting in decision trees,
which may also contribute to improving the accountability of decision trees \cite{ignateiv2021}.

\color{black}
In this study, the input features are assumed to be binary variables when converted to the QUBO problem.
Originally binary variable features such as word occurrences in classical natural language processing \cite{mccallum1998}
or molecular descriptors used
in chemical informatics \cite{mauri2020}, \cite{yap2011} can be used in the proposed method without any pre-processing.
And, even if the variables are not binary variables,
they can be used by using general binarization techniques such as a one-hot encoding \cite{cheng2016recommender}.
Such pre-processing has been done in previous studies \cite{verhaeghe2019}, \cite{verwer2019}.

\section{Related Works}
\textbf{Decision trees}. Decision trees have been used for tabular data analysis in many situations
because of their high performance and accountability \cite{lundberg2018consistent}.
Recent improvements in the performance of decision tree-based methods
such as a random forest \cite{breiman2001random}, XGBoost\cite{chen2016xgboost} and
LightGBM \cite{ke2017lightgbm} have been achieved with techniques for ensembling and boosting decision trees.
\color{black}
However, the method of splitting the samples that make up
the decision tree has not changed significantly
\cite{loh2011classification, grabczewski2005feature}.
In these decision trees, only one feature is used for each splitting using greedy algorithms.
On the other hand, multivariate decision trees use multiple features for splitting
\cite{brodley1995, li2009, murthy1994, kim2001, loh1997, loh1998}.
For example, in \cite{li2009}, linear discriminant function using multiple features is used for splitting.
These studies discussed not only expressive splitting methods but also their efficient search methods.
Furthermore, many studies attempt to optimize the entire decision tree
\cite{narodytska2018, schidler2021, bessiere2009, avellaneda2020}.
These studies aim to reduce the number of splits in the overall decision tree,
which is similar to what this study aims to achieve.

\textbf{Machine learning using MCMC}.
MCMC is an algorithm for sampling from a probability distribution
\cite{metropolis1953equation, hastings1970monte}.
In Bayesian estimation, it is often difficult to derive the posterior distribution computationally.
Therefore, MCMC is commonly used for machine learning
\cite{pinheiro2006mixed,viinikka2020layering,cobb2021scaling}.
Those studies focused on the distributions computed,
whereas we focus on the sampled results by an annealing machine.
Momentum annealing \cite{okuyama2019binary} explores spin configurations in
the ground state by using the Metropolis algorithm \cite{metropolis1953equation},
which is an MCMC method.

\textbf{Annealing machines}.
There has been a lot of research on annealing machines, both in hardware and software.
They have strengths in solving combinatorial optimization problems
and have been used in the real world \cite{stollenwerk2019quantum}.
Quantum annealing \cite{kadowaki1998quantum} and the adiabatic quantum evolution 
algorithm \cite{farhi2001quantum} drove the development of annealing machines.
D-Wave provides a quantum annealing machine in the cloud \cite{shin2014quantum}.
Annealing machines that use GPUs with a technique called momentum annealing
have also been introduced \cite{okuyama2019binary, tao2020}.
The size of the QUBO problem that can be handled differs according to these hardware and algorithm differences,
so it is necessary to select a method that is appropriate for the problem being handled \cite{junger2021}.
As regards software, the library PyQUBO \cite{zaman2021pyqubo} and Ocean \cite{ocean2022} have also been released.

\textbf{Machine learning optimized by annealing machine}.
There have been many attempts to solve problems with annealing technology
by formulating it as a QUBO problem for machine learning algorithms.
In \cite{kurihara2009quantum}, annealing machines have been used for clustering.
In \cite{prasanna2021qubo}, three methods, linear regression, a support vector machine (SVM), and
balanced k-means clustering were transformed into QUBO problems.
There have been also many studies formulating deep learning tasks as QUBO problems
\cite{sasdelli2021quantum, rere2015simulated, sato2019}.
Annealing machines have also been used in image processing
\cite{birdal2021quantum,li2020quantum,cruz2018qubo}, point cloud processing \cite{noormandipour2022, golyanik2020}
and Bayesian network \cite{ogorman2015, shikuri2020}.

\section{Problem Setup}
\label{problemsetup}
Let $x_{r,s_i,r_i}$ be the value of the explanatory raw feature $r_i$,
$t_{s_i}$ be the target variable for sample $s_i$,
and $y_{s_i}$ be the predicted value for sample $s_i$.
Furthermore, let $S$ be the set of all samples,
$S_1$ be the set of samples that satisfied the conditions with split samples,
and $S_0$ be the set of samples that did not.
The number of samples in $S$ is denoted by $N_S$.

\textbf{Binarizing features}.
To easily treat regression trees learning as a QUBO problem,
we binarize the explanatory variable as a preprocessing step.
The binarized features are denoted as $x_{b,s_i,b_i}$ and indicate whether sample $s_i$
satisfies a basic condition $b_i$, as illustrated in Fig. \ref{fig:test4}.
The basic condition is generated from a single raw feature such as $x_{r,s_i,0}>0$.
If sample $s_i$ satisfies the condition $b_i$, then $x_{b,s_i,b_i}=1$; otherwise, $x_{b,s_i,b_i}=0$.
We use logical-product conditions to generate complex conditions for splitting.
A logical-product condition is a condition that is expressed as a logical product of several basic conditions
such as $x_{r,s_i,0}>0 \land x_{r,s_i,1}>0$.
Let $N_F$ and $N_B$ be the numbers of raw feature and basic conditions respectively.

\begin{figure}
  \centering
  \includegraphics[width=1.0\linewidth,page=3]{test4}
  \caption{Binarizing features.
  To easily treat regression trees learning as a QUBO problem,
  the raw explanatory variables are binarized according to the basic conditions.
  There are no restrictions on the binarization method.
  }\label{fig:test4}
\end{figure}

\textbf{Basic conditions}.
The ways to create basic conditions are different for continuous and categorical features.
In the case of continuous features, we make conditions to determine whether a feature is 
both greater than and less than a threshold.
In this way, we can make conditions to specify the value range of a feature, such as $0<x_{r,s_i,0}<1$.
In the case of categorical features, we make conditions to determine
whether a feature is equal to a category variable, such as $x_{r,s_i,0}\neq \text{"}1\text{"}$.
In this way, we can express a logical sum for categorical variables.
For example, if the unique values of the category variables are "1", "2", "3", and "4",
then $x_{r,s_i,0}\neq \text{"}1\text{"} \land x_{r,s_i,0}\neq \text{"}2\text{"}$ represents the same split as 
$x_{r,s_i,0}= \text{"}3\text{"} \lor x_{r,s_i,0}= \text{"}4\text{"}$.
Though we have introduced two typical basic conditions,
other conditions related to whether a feature is missing
and to the relationships between multiple features, such as $x_0>x_1$, are also acceptable.
Note that the only kind of condition that
can be used for splitting samples is a logical-product condition using prepared basic conditions.
In contrast, logical-sum condition is not used.

\textbf{Splitting vector}.
The logical-product condition used for splitting is represented by a binary vector of length $N_B$, 
which is called a splitting vector in this paper.
Each element indicates whether to use each basic condition:
if the value is 1, the condition is used; and if the value is 0, it is not used.
A splitting vector can represent complex conditions via logical-product conditions, as shown in Fig. \ref{fig:test5}.
Because there are $2^{N_B}$ different combinations of conditions,
it is difficult to search for the optimal solution in general.
Therefore, in this paper we use annealing machines,
which have strengths in solving combinatorial optimization problems.
Note that with only one condition to use for splitting,
it is almost the same as the conventional splitting condition search for decision trees
if sufficient basic conditions are prepared.
Specifically, they can be obtained by finding the threshold value of the basic condition
using the inequality on the sorted feature values.
In practice, however, it is not necessary
to use all threshold values in terms of computational both cost and accuracy \cite{ke2017lightgbm} \cite{li2007}.

\begin{figure}
  \centering
  \includegraphics[width=1.0\linewidth,page=3]{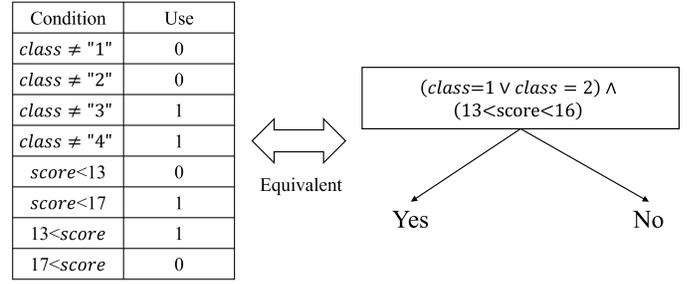}
  \caption{Split representation.
  The conditions used for splitting are represented by binary vectors.
  Each element indicates whether to use a basic condition:
  if the value is 1, the condition is used; and if the value is 0, it is not used.
  }\label{fig:test5}
\end{figure}

\subsection{QUBO Problem}
In an annealing machine, the parameters to be optimized are represented
by spin variables with binary values $\sigma_i\in\{-1,1\}$,
and the objective variable is represented by the Hamiltonian of the Ising model.
The Ising model is given by the following equation:
\begin{equation}
H=-\sum_{i<j}J_{ij}\sigma_i\sigma_j-\sum_ih_i\sigma_i, \label{ising}
\end{equation} 
where $J_{ij}$ is called the spin interaction coefficient, and $h_i$ is the magnetic field coefficient.
The Ising model in Equation \ref{ising} can be transformed by $\sigma_i=2\theta_i-1$ to obtain
the QUBO form:
\begin{equation*}
H=\sum_{i<j}{Q_{ij}\theta_i\theta_j}+\sum_{i}{b_i\theta_i},
\end{equation*}
where the $\theta_i\in\{0,1\}$ are binary variables and called QUBO variables,
and the $Q_{ij}$ and $b_i$ are coefficients that characterize the QUBO problem.
Because the QUBO problem can be expressed in a simpler form than the Ising problem,
in this paper we use the QUBO form.

\subsection{Formulation as QUBO Problem}
The QUBO problem is formulated by expressing the Hamiltonian
as the sum of a loss function $L$ and a constraint function $C$
via the QUBO format, as follow:
\begin{equation*}
  H=L\left(\boldsymbol{X},\boldsymbol{t}|\boldsymbol{\theta}\right)+C(\boldsymbol{\theta})
\end{equation*}
Here, $\boldsymbol{\theta}$ is a binary vector representing the model parameters.
$\boldsymbol{X}$ is the explanatory variable for sample $i$,
and $\boldsymbol{t}$ is a vector representing the objective variable.
In addition, $L$ is the optimization target, and $C(\boldsymbol{\theta})$ is a set to obtain a valid solution,
which is generally required to be $C(\boldsymbol{\theta})=0$.
The annealing machine is used to find $\boldsymbol{\theta}$ such that the Hamiltonian is minimized.

Many constraint functions have been proposed \cite{glover2018tutorial},
and in this paper we use linear inequality constraints.
A linear inequality constraint is one that can be expressed as in Equation \ref{ineq_base},
and the constraint function is represented as in Equation \ref{const_func}:
\begin{gather}
  \alpha\le\sum_{i}a_i\theta_i\le\beta \label{ineq_base}\\
  C(\boldsymbol{\theta})=(\sum_{i}a_i\theta_i-\sum_{\alpha\le j \le \beta}j\theta_{s,j})^2+
  (\sum_{\alpha\le j \le \beta}\theta_{s,j}-1)^2, \label{const_func}
\end{gather}
where $\theta_{s,j}$ is an auxiliary variable and is the target of optimization.
The constraints of Equation \ref{ineq_base} are satisfied when $C(\boldsymbol{\theta})=0$.
In the case where $\alpha$ and $\beta$ are equal, which corresponds to the equality constraint,
the second term in Equation \ref{const_func} is unnecessary.
\subsection{Contributions}
Our contributions in this paper are as follows.
\begin{enumerate}
  \item We propose a method to formulate the search problem of finding splitting conditions for regression trees in 
  a QUBO problem that can be solved by annealing machines.
  \item We extend regression trees to search more complicated conditions,
  namely, splitting conditions defined by multi-dimensional boundaries.
  \item To improve the solution's convergence,
  we propose a splitting constraint, which 
  requires the existence of both samples that satisfy the splitting condition and samples that do not.
  \item We evaluate the proposed method on both synthetic and real data.
  As a result, although there are restrictions on the size of the problem that can be handled,
  we show that it is possible to stochastically find complex conditions on synthetic data and
  conditions to reduce the estimation error compared conventional splitting method.  
\end{enumerate}
Overall, in this paper we show that regression trees, which are widely used in tabular data analysis,
can be made more sophisticated by using annealing machines.
This technique provides a new option for machine learning.

\section{QUBO Decision Tree}
We first convert the error function, the mean squared error (MSE), to the QUBO format.
The MSE in a regression tree can be written as follows:
\begin{equation*}
  \text{MSE}=\frac{1}{N_S}(\sum_{{s_i}\in S_1}\left({\text{pred}}_1-t_{s_i}\right)^2+
  \sum_{{s_i}\in S_0}\left({\text{pred}}_0-t_{s_i}\right)^2),
\end{equation*}
where $\text{pred}_1$ is the estimated value when the condition is satisfied,
and $\text{pred}_0$ is the estimated value when it is not satisfied.
Learning in regression trees involves finding the conditions for splitting
and the $\text{pred}_1,\text{pred}_0$ to reduce the MSE.
It is obvious that $\text{pred}_1$ and $\text{pred}_0$ are the means of the samples divided by the condition.
Therefore, the MSE is equal to sum of the variance of the split sample groups
weighted by the proportions of the sample groups.
Accordingly, the MSE can be written as the following equation:
\begin{align*}
  \text{MSE}=&\sum_{b=0,1}{\text{Var}\left(\left\{t_{s_i}\middle| s_i\in S_b\right\}\right)\frac{N_{S_b}}{N_S}}\\
  =&\sum_{b=0,1}{(\frac{1}{N_S}\sum_{s_i\in S_b} t_{s_i}^2-\frac{1}{N_SN_{S_b}}(\sum_{s_i\in S_b} t_{s_i})^2)}.
\end{align*}
However, the MSE is difficult to formulate in a QUBO problem because
it involves division using the variable $N_{S_b}$ that is the sum of QUBO variables.
Here, because $N_S$ is the total number of samples and thus a constant,
this is not a problem.
Hence, we propose the square weighted MSE (SWMSE).
In the SWMSE, instead of using the proportion of the sample group as the weight in calculating the sum of the variances,
the square of that value is used:
\begin{align}
  \text{SWMSE}=&\sum_{b=0,1}{\text{Var}\left(\left\{t_{s_i}\middle| s_i\in S_b\right\}\right)\left(\frac{N_{S_b}}{N_S}\right)^2} \notag \\
  =&\sum_{b=0,1}{(\frac{N_{S_b}}{N_S}\sum_{s_i\in S_b} t_{s_i}^2-\frac{1}{N_S}(\sum_{s_i\in S_b} t_{s_i})^2)}. \label{SWMSE}
\end{align}
Through this transformation, the MSE minimization problem becomes a QUBO problem.
The relationship between the MSE and SWMSE is discussed in section \ref{sec:discussion}.
\subsection{Formulation as QUBO Problem}
There are two types of QUBO variables to be optimized $\boldsymbol{\theta}_B$ and $\boldsymbol{\theta}_X$,
and their elements are denoted by $\theta_{B,b_i}$ and $\theta_{X,s_i,c_i}$, respectively.
Let $\boldsymbol{\Theta}_B$ be the set of possible vectors of $\boldsymbol{\theta}_B$.
First, $\theta_{B,b_i}$ is a binary variable that represent the splitting logical-product condition.
The samples are split according to whether they satisfy the logical product of 
the conditions $b_i$ that satisfy $\theta_{B,b_i}=1$.
$\theta_{X,s_i,c_i}$ is an auxiliary binary variable for calculating the SWMSE,
where $\theta_{X,s_i,c_i}=1$ implies that
there are $c_i$ conditions that sample $s_i$ does not satisfy.
Note that $\theta_{X,s_i,0}=1$ indicates that sample $s_i$ satisfies all the splitting conditions,
i.e., $\theta_{X,s_i,0}$ indicates the splitting result.
Here, $c_i$ is an integer from 0 to $M$,
where $M$ is the maximum number of binary features to be used in the logical-product conditions,
and is a learning parameter.
The number of samples $s_i$ that do not satisfying the conditions,
denoted as $unsatisfied\_feature_{s_i}$,
is expressed by the following equation:
\begin{equation*}
  unsatisfied\_feature_{s_i}=\sum_{b_i}\left(1-x_{b,s_i,b_i}\right)\theta_{B,b_i}.
\end{equation*}
If $unsatisfied\_feature_{s_i}=0$, then the sample $s_i$ satisfies all the splitting condition.

\textbf{Loss function}.
The moment of $t_{s_i}$ appearing in Equation \eqref{SWMSE} can be expressed
by using QUBO variables as in the following equations:
\begin{align*}
  \sum_{s_i\in S_1} t_{s_i}^n &= \sum_{s_i} \theta_{X,s_i,0}t_{s_i}^n \;\;\;\;\;\;\;\;\;\;\;(n=1,2,..),\\
  \sum_{s_i\in S_0} t_{s_i}^n &= \sum_{s_i} (1-\theta_{X,s_i,0})t_{s_i}^n \;\;\;(n=1,2,..).
\end{align*}
Similarly, the numbers of sample groups, $N_{S_1}$ and $N_{S_0}$, that are split by a condition can be written as follows:
\begin{align*}
  N_{S_1} &= \sum_{s_i} \theta_{X,s_i,0}\\
  N_{S_0} &= \sum_{s_i} (1-\theta_{X,s_i,0}).
\end{align*}
As a result, the SWMSE can be expressed as a problem in QUBO form
via the following equation:
\normalsize
\begin{align*}
  \text{SWMSE}&=(\sum_{s_i}\theta_{X,s_i,0} t_{s_i}^2)(\sum_{s_i}\theta_{X,s_i,0})\\
  &-(\sum_{s_i}\theta_{X,s_i,0} t_{s_i})^2\\
  &+(\sum_{s_i}(1-\theta_{X,s_i,0}) t_{s_i}^2)(\sum_{s_i}(1-\theta_{X,s_i,0}))\\
  &-(\sum_{s_i}(1-\theta_{X,s_i,0} t_{s_i}))^2.
\end{align*}
\normalsize
\textbf{Constraint function}.
There are three types of constraints on QUBO variables,
as represented by Equation \eqref{cnstr1}-\eqref{cnstr3} below.
Here, Equation \eqref{cnstr1} is a constraint on the relationship between
$\theta_{B,b_i}$ and $\theta_{X,s_i,c_i}$ for each sample.
Satisfaction of this constraint indicates that $\theta_{X,s_i,c_i}$ can represent
the number of satisfied basic conditions that constitute the splitting condition in each sample.
Equation \eqref{cnstr2} is a constraint on the validity of $\theta_{X,s_i,c_i}$ for each sample.
Equation \eqref{cnstr3} is an optional constraint for narrowing down the search space.
Though $M$ can take values up to $N_B$, we can limit it for practical purposes.
\begin{gather}
  \forall s_i \sum_{b_i} (1-x_{b,s_i,b_i})\theta_{B,b_i}-\sum_{c_i}c_i\theta_{X,s_i,c_i}=0 \label{cnstr1}\\
  \forall s_i \sum_{c_i} \theta_{X,s_i,c_i}=1 \label{cnstr2}\\
  1\leq\sum_{b_i} \theta_{B,b_i} \leq M \label{cnstr3}
\end{gather}
When all these constraints are satisfied,
the actual SWMSE is equal to the SWMSE being calculated in the annealing machine.
Conversely, because, when a few constraints are violated,
the difference between the two SWMSE is small, and the impact on learning is also small,
not all constraints must be satisfied.

\textbf{Splitting constraint}.
In addition to the three constraints above, we also propose a constraint for making an effective partition.
By adding this constraint,
we can prevent the number of samples after splitting from being zero, which stabilizes the learning.
Here, $a$ is the minimum sample ratio after splitting.
\begin{equation}
aN_S \le \sum_{s_i}\theta_{X,s_i,0} \le (1-a)N_S \label{cnstr4}
\end{equation}
\textbf{Hamiltonian}.
The splitting method is explored by minimizing the Hamiltonian given below in Equation \eqref{finalH}.
In this equation, $C_1,C_2,C_3$, and $C_{add}$ are the Hamiltonians for
the constraints expressed in Equations \eqref{cnstr1}-\eqref{cnstr4},
and $w$ is a parameter that adjusts the weight of each Hamiltonian.
Note that not all of these constraints must be satisfied.
In other words, even if there is a violated constraint,
an estimator can be created using obtained $\theta_{B,b_i}$.
\begin{align}
  H&=\frac{w_q}{N_S}\text{SWMSE}(\boldsymbol{X},\boldsymbol{t}|\boldsymbol{\theta}_B,\boldsymbol{\theta}_X)
  +\frac{w_{c1}}{N_S}C_1(\boldsymbol{\theta}_B,\boldsymbol{\theta}_X)\notag \\
  &+\frac{w_{c2}}{N_S}C_2(\boldsymbol{\theta}_X)+C_3(\boldsymbol{\theta}_X)
  +C_{add}(\boldsymbol{\theta}_X) \label {finalH}
\end{align}
\textbf{Prediction}.
An estimator is created according to the $\theta_{B,b_i}$ obtained from the annealing machine, 
as represented by the following equations.
\begin{gather*}
  S_1 = \{s_i|s_i \in S \land \sum_{b_i}\left(1-x_{b,s_i,b_i}\right)\theta_{B,b_i}=0\}\\
  S_0 = \{s_i|s_i \in S \land \sum_{b_i}\left(1-x_{b,s_i,b_i}\right)\theta_{B,b_i}>0\}\\
  y_{s_i} = 
  \begin{cases}
    \frac{1}{N_{S_1}}\sum_{s_i \in S_1} t_{s_i}, & s_i \in S_1 \\
    \frac{1}{N_{S_0}}\sum_{s_i \in S_0} t_{s_i}, & s_i \in S_0
\end{cases}
\end{gather*}

\section{Experiments}
Simulated Annealing, because of its stochastic process, returns a different solution for each trial.
Accordingly, the optimization is carried out multiple times with different random seeds,
and the final result is determined by comparing the obtained results.
In this study, we evaluated the proposed method
by observing numbers of desirable results that were obtained in multiple trials.

\textbf{Synthetic datasets}. 
To evaluate the proposed method's capability to find the optimal conditions,
we generated synthetic datasets.
These datasets were generated using three parameters: $K,N_S$, and $N_F$, as illustrated in Fig. \ref{fig:test14}.
Here, $K$ is the number of features that make up the logical-product condition, 
and the objective variable $t_{s_i}$ is determined from
$x_{b,s_i,0},x_{b,s_i,1},..,x_{b,s_i,K-1}$ via the following equation:
\begin{gather*}
  t_{s_i} = 
  \begin{cases}
    1 \;\;\;\text{if}\;\;x_{b,s_i,0}=x_{b,s_i,1}=..=x_{b,s_i,K-1}=1\\
    0 \;\;\;\text{else}
  \end{cases}.
\end{gather*}
Because $x_{b,s_i,b_i}$ is a binary variable, we can also write {$t_{s_i}=\prod_{b_i=0}^{K-1}x_{s_i,b_i}$.
It is also clear that $\boldsymbol{\theta_B}$ satisfying
$\theta_{B,0}=1,..,\theta_{B,K-1}=1,\theta_{B,K}=0,..\theta_{B,N_B-1}=0$
is the optimal splitting vector $\boldsymbol{\theta^*_B}$ that achieves the minimum MSE (Fig. \ref{fig:test14}).
In terms of spam message classification using word occurrence vectors,
it is possible to split messages that contain all the selected words from those that do not.
Specifically, the dataset generation was based on algorithm \ref{alg1},
whose parameters were adjusted so that the average of $\boldsymbol{t}$ was 0.5.

\begin{figure}
  \centering
  \includegraphics[width=1.0\linewidth,page=3]{test14}
  \caption{Synthetic datasets.
  The datasets were generated using three parameters: $K,N_S$, and $N_F$.
  If $x_{b,s_i,0}=x_{b,s_i,1}=..=x_{b,s_i,K-1}$ is 1,
  then the objective variable $t_{s_i}=1$.
  $\boldsymbol{\theta^*_B}$ is the optimal splitting vector.
  }\label{fig:test14}
\end{figure}

\begin{figure}[!t]
  \begin{algorithm}[H]
      \caption{Generate Synthetic Data}
      \label{alg1}
      \begin{algorithmic}[1]    
      \renewcommand{\algorithmicrequire}{\textbf{Input:}}
      \renewcommand{\algorithmicensure}{\textbf{Output:}}
      \REQUIRE $N_S,N_B,K$
      \ENSURE $x_{s_i,b_i},t_{s_i}$
      \FOR {$s_i=0$ to $N_B-1$}
      \FOR {$b_i=0$ to $N_B$}
      \STATE $rand \leftarrow Uniform(0,1)$
      \IF{$b_i < K$}
      \STATE $th \leftarrow (1-0.5^{1/K})$
      \ELSE
      \STATE $th \leftarrow 0.5$
      \ENDIF
      \IF {$rand>th$}
      \STATE $x_{s_i,b_i}=1$
      \ELSE
      \STATE $x_{s_i,b_i}=0$
      \ENDIF
      \ENDFOR
      \STATE{$t_{s_i}=\prod_{b_i=0}^{K-1}x_{s_i,b_i}$}
      \ENDFOR
      \end{algorithmic}
  \end{algorithm}
  \end{figure}

\textbf{Real datasets}. To evaluate the proposed method's capability to
train a regression tree on real data,
we used the Ames Housing dataset \cite{de2011ames}.
This dataset contains data for 1460 housed on 79 explanatory variables and the objective variable,
house prices.
To binarize the features, continuous variables were divided by $q$-quantiles ($q=0.33,0.66$),
and categorical features with unique values above 3 were removed.
As a result, the final number of binary features was 136.
\subsection{Evaluation}
\textbf{Synthetic datasets}.
Using the synthetic datasets, we evaluated the proposed method
in term of its ability to find the optimal splitting vector $\boldsymbol{\theta^*_B}$,
which satisfies $\theta^*_{B,0}=1,..,\theta^*_{B,K-1}=1,\theta^*_{B,K}=0,..\theta^*_{B,N_B-1}=0$.
These experiments were performed on a CPU using the library PyQUBO.
The parameter settings were 1000 for the number of trials and
10000 for the number of iterations in each trial, with default values for the other parameters.
We evaluated the number of trials in which the optimal splitting vector was found.
Five experiments were conducted,
and the datasets were generated randomly.
The mean and standard deviation of the number of trials
to find the optimal splitting vector were calculated.

Fig. \ref{fig:test1} shows the experimental results with $K=1$, $M=1$, 
and various values of $N_S (=20,50,100)$ and $N_B (=10,50,100)$.
As a result, we found that as the problem size increased,
the number of trials to find the optimal splitting vector decreased.
The number of features that affected the number of conditions, $N_B$, had an especially large impact on the result.
These results demonstrate that the same splitting result obtained by the decision tree in the conventional method
can be obtained stochastically by using the proposed method.
\begin{figure}
  \centering
  \includegraphics[width=1.0\linewidth,page=3]{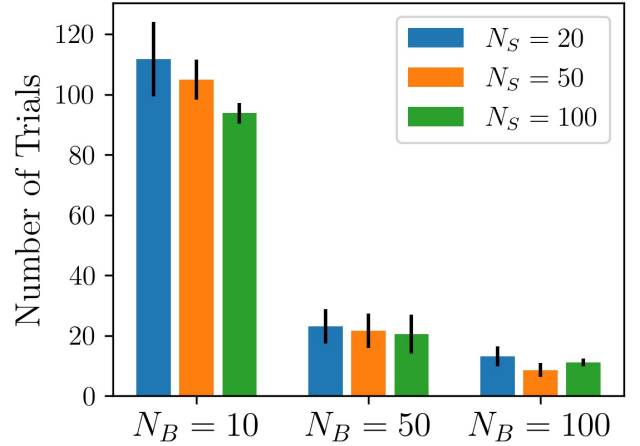}
  \caption{The Number of trials to find the optimal splitting vector ($K=1$).
  The black lines indicate the standard deviations.
  }\label{fig:test1}
\end{figure}

Table \ref{tab:data1} lists the experimental results with $K=1, N_S=20$, $N_B=10$, 
and various values of $M (=1,2,3,4)$.
The number of trials to find the optimal splitting vector decreased as $M$ increased,
and when $M$ was 4, the number of trials was 0 for all $N_S$.
We expect that the number of trials to find the optimal splitting decreases
as more $\boldsymbol{\theta}_B$ satisfy the constraints.

\begin{table}
  \centering
  \caption{Relationship between $M$ and the number of trials to find the optimal splitting vector.
  }\label{tab:data1}
  \begin{tabular}{|c|c|c|c|c|}
     \hline
     &M=1&M=2&M=3&M=4 \\
     \hline
    $N_S$=20&111.8$\pm$12.3 & 69.4$\pm$20&2.8$\pm$2.5&0.0$\pm$0.0\\
    $N_S$=50&105.0$\pm$6.6 &5.8$\pm$5.2&0.0$\pm$0.0&0.0$\pm$0.0\\
    $N_S$=100&93.8$\pm$3.3 &0.8$\pm$0.8&0.0$\pm$0.0&0.0$\pm$0.0\\
    \hline
  \end{tabular}
\end{table}

Next, we evaluated whether the proposed method could find the optimal splitting vector
that represented logical-product condition.
This was verified by using the data generation algorithm with $K=2$,
in which case $\theta_{B,b_i}=\{1,1,0,...,0\}$ is the optimal splitting vector.
Table \ref{tab:data2-1} lists the numbers of trials that found the optimal splitting vector for $M=2$ and $K=1,2$.
Even when the conditions were made more complicated, no extreme accuracy degradation was observed,
which confirmed that the search space of $\theta_{B,b_i}$ had a significant impact.

\begin{table}
  \centering
  \caption{Relationship between $K$ and the number of trials to find the optimal splitting vector.
  }\label{tab:data2-1}                                          
  \begin{tabular}{|c|c|c|}
     \hline
     &K=1&K=2\\
     \hline
    $N_S$=20& 69.4$\pm$20&61.0$\pm$14.0\\
    $N_S$=50&5.8$\pm$5.2&8.2$\pm$3.6\\
    $N_S$=100&0.8$\pm$0.8&2.4$\pm$1.3\\
    \hline
  \end{tabular}
\end{table}

\textbf{Real dataset}.
Using the real dataset, we evaluated the proposed method's capability
to find a splitting vector achieving a low MSE.
We denote the MSE obtained with a splitting vector of $\boldsymbol{\theta}_B$ as $\text{MSE}(\boldsymbol{\theta}_B)$.
We evaluated whether the proposed method could find a splitting vector that achieved
an MSE that was lower than the best MSE when only one basic condition was used.
The best MSE with only one basic condition is denoted as cMSE and was obtained via the following equations:
\begin{gather*}
  \boldsymbol{\Theta}^1_B = \{\boldsymbol{\theta}_B|\boldsymbol{\theta}_B\in \boldsymbol{\Theta}_B
  \land \sum_{b_i}\theta_{B,b_i}=1\},\\
  \text{cMSE}=\min_{\boldsymbol{\theta}_B \in \boldsymbol{\Theta}^1_B}{\text{MSE}(\boldsymbol{\theta}_B)},
\end{gather*}
where $\boldsymbol{\Theta}^1_B$ is the set of splitting vectors with only one condition.
If the basic conditions are sufficiently prepared, cMSE is consistent with
the MSE in using conventional regression trees.
The experiments on the real dataset were performed on a GPU using momentum annealing.
The parameter of the number of trials was set to 1000.
Ten experiments were conducted,
and the samples were selected randomly form the dataset.
The mean and standard deviation of the number of trials were again calculated.
We set the minimum split ratio $a=0.2$.

\begin{figure}
  \centering
  \includegraphics[width=1.0\linewidth,page=3]{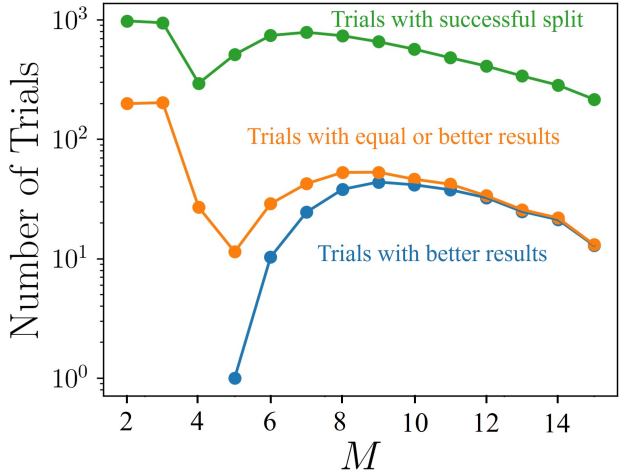}
  \caption{Results of evaluating the splits with different $M$ values.
  Note that there were no trials with better results for $M=2,3,4$.}\label{fig:test9}
\end{figure}

\begin{figure}
  \centering
  \includegraphics[width=1.0\linewidth,page=3]{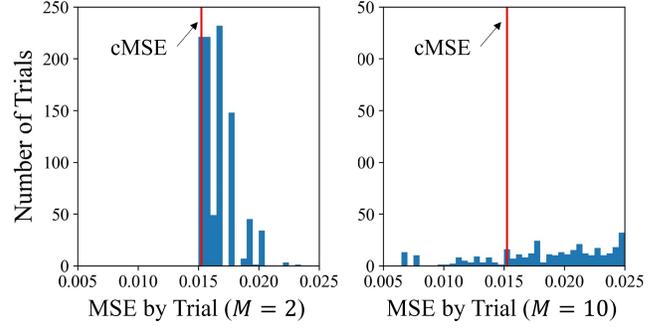}
  \caption{Histograms of the MSE values.
  The red lines indicate the cMSE values.}\label{fig:test15}
\end{figure}

Fig. \ref{fig:test9} shows the results of evaluating each trial ($N_S=20$) with different $M$ values
from three perspectives: whether splittable conditions were generated,
whether an MSE equal or superior to that of the comparison method
was achieved ($\text{MSE}(\boldsymbol{\theta}_B)\leq\text{cMSE}$),
and whether an MSE superior to the comparison method was achieved ($\text{MSE}(\boldsymbol{\theta}_B)<\text{cMSE}$).
Here a splittable condition means a condition for which there exist
both samples that satisfy the condition and samples that do not.
Let $N_{gs}$, $N_{eq}$, and $N_{su}$ be the numbers of trials that satisfied each of the respective conditions.
Interestingly, $N_{gs}$ and $N_{eq}$ decreased as $M$ increased.
However, $N_{su}$ had its maximum value at $M=9$.
This may have been due to convergences to a local solution with a strong constraint on $\theta_{B,b_i}$.

Fig. \ref{fig:test15} shows histograms of the MSE values for $M=2,10$.
When $M$ was 2, many trials were concentrated around the MSE value for the conventional method.
On the other hand, when $M$ was 10, the MSE values were widely distributed,
and in several trials, MSE was lower than that of the conventional method.
Table \ref{tab:data2} lists the $N_{su}$ for various values of $N_S$ and $M$.
Even when the number of samples increased,
the proposed method could not find a better split with a small $M$.
In Table \ref{tab:data3}, we summarize the effect of the splitting constraint.
It can be seen that the addition of the splitting constraint increased $N_{su}$ on average.

\begin{table}
  \centering
  \caption{$N_{su}$ with different $N_S$ and $M$ values.}\label{tab:data2}
  \begin{tabular}{|c|cccc|}
    \hline
     $N_S$&3&5&8&10 \\
     \hline
    20 & 0.0$\pm$0.0 & 1.0$\pm$0.9&38.0$\pm$19.8&41.5$\pm$29.1\\
    30 & 0.0$\pm$0.0 & 3.4$\pm$1.7&2.2$\pm$1.8&1.3$\pm$1.6\\
    40 & 0.0$\pm$0.0 & 0.4$\pm$0.7&0.0$\pm$0.0&0.0$\pm$0.0\\
    50 & 0.0$\pm$0.0 & 0.0$\pm$0.0 & 0.0$\pm$0.0 & 0.0$\pm$0.0 \\
    \hline
  \end{tabular}
\end{table}

\begin{table}
  \centering
  \caption{Ablation results on the effect of splitting constraints.}\label{tab:data3}
  \begin{tabular}{|c|cccc|}
\hline
     &3&5&8&10 \\
\hline
    w/o & 0.0$\pm$0.0 & 0.8$\pm$1.1&22.7$\pm$12.7&22.7$\pm$17.1\\
    w/ & 0.0$\pm$0.0 & 1.0$\pm$0.9&38.0$\pm$19.8&41.5$\pm$29.1\\
\hline
  \end{tabular}
\end{table}

\textbf{Condition redundancy}.
Some of the splitting conditions that made up the obtained logical-product condition
did not affect the splitting result.
Such redundant conditions may have had a negative impact during the evaluation.
Accordingly, we removed conditions from the logical-product condition
that did not change the splitting result until no such conditions remained.
Note that this process does not guarantee the minimum number of conditions to achieve the split.
Fig. \ref{fig:test10} shows the number of features used before and after this redundancy removal 
process ($M=10,N_S=20$).
To improve the data's visibility, noise is added to the coordinate positions of each point.
As shown in the figure, only about half of the conditions were actually valid.
We expect that these redundant splitting conditions contributed to 
preventing the converges to a local solution.
More importantly, we learned that we need to tune the parameter $M$ well when we use it.

\begin{figure}
  \centering
  \includegraphics[width=1.0\linewidth,page=3]{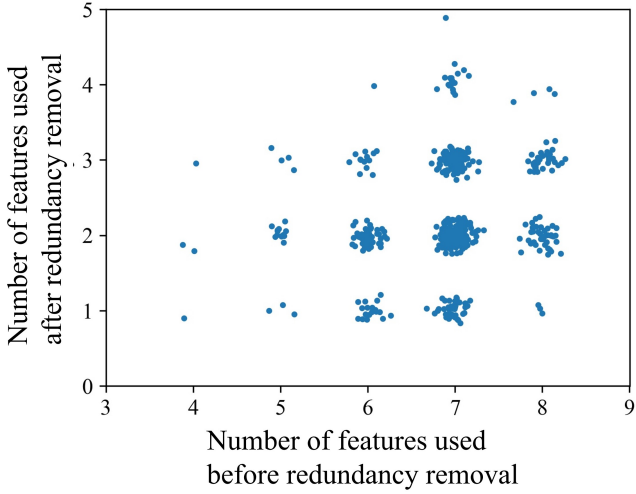}
  \caption{The Numbers of features used before and after redundancy removal 
  ($M=10,N_S=20$), with noise added to improve visibility.
  }\label{fig:test10}
\end{figure}

\section{Discussion}\label{sec:discussion}

\textbf{Relationship between MSE and SWMSE}. 
We first discuss the differences between the MSE and SWMSE.
Let $\rho$ and $V$ be the proportion of sample groups that satisfy the condition and the variance of the objective variable,
respectively, and let $\gamma V$ be the proportion of sample groups that do not satisfy the condition.
Then, the two indicators can be written as follows.
\begin{align*}
  \text{SWMSE}&=V\rho^2+\gamma^2V(1-\rho)^2\\
  &=V(\rho^2+\gamma^2(1-\rho)^2))\\
  &=(\frac{\text{MSE}}{\rho+\gamma^2(1-\rho)}) (\rho^2+\gamma^2(1-\rho)^2)\\
  &= \frac{\rho^2+\gamma^2 (1-\rho)^2}{\rho+\gamma^2 (1-\rho)} \cdot \text{MSE}
\end{align*}
Fig. \ref{fig:test11} shows the maximum and minimum values of the ratio of the two indices
when the minimum split ratio $a$ using the splitting constraint was varied,
assuming $0.5<\gamma<2.0$.
As $a$ increased, the ratio's range decreased,
suggests that the splitting constraint contributed to mitigating the difference between
the SWMSE and the MSE in the optimization.

\begin{figure}
  \centering
  \includegraphics[width=1.0
  \linewidth,page=3]{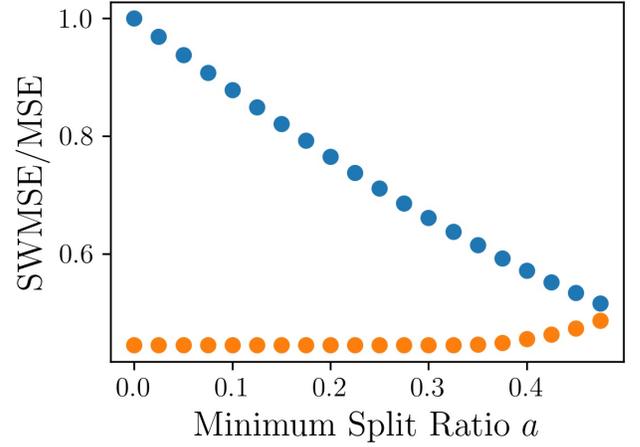}
  \caption{Relation between the minimum split ratio and the SWMSE/MSE ratio.
  The blue and orange points represent the maximum and minimum values,
  respectively of the SWMSE/MSE.}\label{fig:test11}
\end{figure}

Furthermore, assuming that the variances of the $y$ values of the sample groups after splitting are equal,
the following equation holds.
\begin{equation*}
  \text{SWMSE}= (2(\rho_1-0.5)^2+0.5)\cdot\text{MSE}
\end{equation*}
If the MSE is the same, the SWMSE works as the samples are equally divided.
This may contribute to preventing overfitting,
and the difference between the two indices does not seem to have an extremely negative impact on learning.

\textbf{Differences from regression trees with depth}:
There are two advantages in using a shallow logical-product condition.
First, it does not extra splitting of samples that do not satisfy the condition.
In regression trees with depth, samples that do not satisfy the condition are split into different groups.
As a result, fewer samples can be used for predictions, which makes them more vulnerable to noise.
Second, the existence of many splits reduces the model's explainability.
In fact, it is rarely possible to see the regression tree split one by one.

\textbf{Actual use}.
As shown in our experiments,
the process of finding a good split is stochastic, and in some cases, less accurate than conventional methods.
To overcome this problem, 
it would be better to use both the proposed method and deterministic conventional methods,
and to then select the more accurate method for each splitting.
In addition, as the proposed method is concerned with splitting,
it seems not to be difficult to apply techniques related to decision trees such as ensembles, boosting and early stopping.

\section{Conclusion}
In this paper, we formulated the search problem of finding a splitting to reduce the prediction error
in regression trees as a QUBO problem.
The QUBO formulation made it possible to search for complex splitting conditions using multiple features.
We showed the effectiveness of the proposed splitting method.
On synthetic data, we showed The ability of the proposed method to discover conditions that can be expressed as logical products.
On real data, we showed The ability of the proposed method to discover better conditions for splitting than conventional methods.
However, in an actual decision tree, multiple splits are repeated to output the final estimation.
In the future, the regression tree techniques developed so far will need to be applied to the proposed method and verified.
We believe that these techniques will provide a new option for machine learning.




\appendix
\subsection{Reproducibility}
The entire generation process of the synthetic data used in this study data is described, 
and the synthetic data is reproducible except for probalistic elements.
The real data is publicly available \cite{de2011ames}.
With regard to the source codes,
those using artificial data provide a description of the parameters sufficient to be calculated using PyQUBO \cite{zaman2021pyqubo}.

\subsection{Number of Binarized Features}
This section specifically describes the method of feature binarization and explains the final number of binarized features.
First, assume that all features are given as either continuous or categorical variables.
If they are not given, all features are treated as continuous variables.
In the case of categorical features, we make conditions to determine
whether a feature is equal to a category variable.
In other words, as many binary features are created as there are unique feature values.
In the case of continuous features, we make conditions to determine whether a feature is 
both greater than and less than a threshold.
The quantile is a way of determining these thresholds.
Setting $1/Q, 2/Q\dots,(Q-1)/Q$ with parameter $Q$ generate  a total of $2(Q-1)$ binary features.
If all features were continuous, it could be written as follows.
\begin{equation*}
 N_B = 2N_R(Q-1)
\end{equation*}
The maximum possible value of Q is $N_S-1$, $N_B$ can be written as follows.
\begin{equation*}
  N_B \leq 2N_R(N_S-2)
 \end{equation*}
 Binarization in this way results in a large number of QUBO variables,
 so $Q$ needs to be set appropriately.
 As mentioned in section \ref{problemsetup}, reducing $Q$ appropriately is effective
 not only in terms of computational cost but also in terms of accuracy.

 \begin{table}[h]
  \centering
  \caption{Number of QUBO variables.}\label{sup}
  \begin{tabular}{|c|c|}
    \hline
     &Number of QUBO variables \\
     \hline
    $\theta_{B,b_i}$ & $N_B(\lneq 2N_R(N_S-2))$\\
    $\theta_{X,s_i,c_i}$ & $N_S(M+1)$\\
    $\theta_{s,c3,j}$&$M$\\
    $\theta_{s,add,j}$&$N_S(1-2a)$\\
    \hline
  \end{tabular}
\end{table}

\subsection{Final Hamiltonian and Number of QUBO Variables}
The final Hamiltonian in the proposed method is as follows.
\begin{align*}
  H=&\frac{w_q}{N_S}\text{SWMSE}(\boldsymbol{X},\boldsymbol{t}|\boldsymbol{\theta}_B,\boldsymbol{\theta}_X)\\
  &+\frac{w_{c1}}{N_S}C_1(\boldsymbol{\theta}_B,\boldsymbol{\theta}_X)\notag
  +\frac{w_{c2}}{N_S}C_2(\boldsymbol{\theta}_X)+C_3(\boldsymbol{\theta}_X)
  +C_{add}(\boldsymbol{\theta}_X)\\
  =&\frac{w_q}{N_S}\left((\sum_{s_i}\theta_{X,s_i,0} t_{s_i}^2)(\sum_{s_i}\theta_{X,s_i,0})
  -(\sum_{s_i}\theta_{X,s_i,0} t_{s_i})^2 \right. \\
  &+(\sum_{s_i}(1-\theta_{X,s_i,0}) t_{s_i}^2)(\sum_{s_i}(1-\theta_{X,s_i,0})) \\
  &\left.-(\sum_{s_i}(1-\theta_{X,s_i,0} t_{s_i}))^2\right)\\
  &+\frac{w_{c1}}{N_S}\left(\sum_{s_i}\left(\sum_{b_i} (1-x_{b,s_i,b_i})\theta_{B,b_i}-
  \sum_{c_i}c_i\theta_{X,s_i,c_i}\right)^2\right)\\
  &+\frac{w_{c2}}{N_S}\left(\sum_{s_i}\left(\sum_{c_i} \theta_{X,s_i,c_i}-1\right)^2\right)\\
  &+\left((\sum_{b_i}\theta_{B,b_i}-\sum_{1\le j \le M}j\theta_{s,c3,j})^2+
  (\sum_{1\le j \le M}\theta_{s,c3,j}-1)^2\right)\\
  &+\left(\sum_{s_i}\theta_{X,s_i,0}-\sum_{aN_S\le j \le (1-a)N_S}j\theta_{s,add,j})^2\right.\\
  &+\left.(\sum_{aN_S\le j \le (1-a)N_S}\theta_{s,add,j}-1)^2\right)
\end{align*}
In general, there is a limit to the number of QUBO variables that annealing machines can handle.
The number of QUBO variables in the QUBO problem created by the proposed method is
$N_B+N_S(M+1)+M+N_S(1-2a)$, as listed in Table \ref{sup}.
Because $N_S$, $N_B$, and $M$ have a great influence on the number of QUBO variables,
these values have to be adjusted according to the task and the hardware.
In this formulation there are interactions between many pairs of QUBO variables.
Therefore, it is better to use annealing machines that can handle fully connected QUBO problems easily.
If it is difficult to deal with this problem directly,
it is necessary to transform the problem using graph-minor embedding techniques
and to ensure that the QUBO variables do not increase too much \cite{venturclli2015}.

\vspace{12pt}


\begin{thebibliography}{00}
  \bibitem{martovnak2004quantum} R. Martok, G. E. Santoro and E. Tosatti, "Quantum annealing of the traveling-salesman problem", Phys. Rev. E Stat. Phys. Plasmas Fluids Relat. Interdiscip. Top., vol. 70, no. 5, pp. 057701, Nov. 2004.
  \bibitem{shin2014quantum} S. W. Shin, G. Smith, J. A. Smolin and U. Vazirani, "How “quantum” is the d-wave machine?", arXiv preprint arXiv:1401.7087, 2014.
  \bibitem{ibm2021} "IBM Quanum" 2022, [Online]. Available: https://quantum-computing.ibm.com/.  [Accessed on the 08.06.2022].
  \bibitem{kadowaki1998quantum} T. Kadowaki and H. Nishimori, "Quantum annealing in the transverse Ising model", Phys. Rev. E, vol. 58, pp. 5355-5363, Nov 1998.
  \bibitem{Justin1999} J. J. Brooke, D. Bitko, T. F. Rosenbaum, and G. Aeppli. Quantum annealing of a disordered magnet. Science, 284(5415):779–781, 1999.
  \bibitem{prasanna2021qubo} P. Date, D. Arthur and P. Lauren, "Qubo formulations for training machine learning models", Scientific Reports, vol. 11, no. 1, pp. 1-10, 2021.
  \bibitem{safavian1991survey} S. R. Safavian and D. Landgrebe, "A survey of decision tree classifier methodology", in IEEE Transactions on Systems, Man, and Cybernetics, vol. 21, no. 3, pp. 660-674, May-June 1991, doi: 10.1109/21.97458.
  \bibitem{chen2016xgboost} T. Chen and C. Guestrin, "XGBoost: A scalable tree boosting system", Proc. SIGKDD, pp. 785-794, 2016.
  \bibitem{ke2017lightgbm} G. Ke et al., "LightGBM: A highly efficient gradient boosting decision tree", Proc. 31st Conf. Neural Inf. Process. Syst. (NIPS), pp. 3146-3154, 2017.
  \bibitem{Bennett2007} J. Bennett and S. Lanning. The netflix prize. In Proceedings of the KDD Cup Workshop 2007, pages 3–6, New York, Aug. 2007.
  \bibitem{He2014} X. He, J. Pan, O. Jin, T. Xu, B. Liu, T. Xu, Y. Shi, A. Atallah, R. Herbrich, S. Bowers, and J. Q. n. Candela. Practical lessons from predicting clicks on ads at facebook. In Proceedings of the Eighth International Workshop on Data Mining for Online Advertising, ADKDD’14, 2014.
  \bibitem{breiman2001random} L. Breiman. Random forests. Technical report, 2001
  \bibitem{loh2011classification} W. Y. Loh, "Classification and regression trees", Wiley Interdisciplinary Rev. Data Mining Knowl. Discovery, vol. 1, pp. 14-23, 2011.
  \bibitem{grabczewski2005feature} K. Grabczewski and N. Jankowski, "Feature selection with decision tree criterion," Fifth International Conference on Hybrid Intelligent Systems (HIS'05), 2005, pp. 6 pp.-, doi: 10.1109/ICHIS.2005.43.
  \bibitem{kirkpatrick1983optimization} S. A. Kirkpatrick, C. D. Gellatt Jr. and M. P. Vecchi, "Optimization by simulated annealing", Science, vol. 220, pp. 671-680, 1983.
  \bibitem{ignateiv2021} A. Ignateiv, et al. A scalable two stage approach to computing optimal decision sets. arXiv preprint arXiv:2102.01904, 2021.
  \bibitem{mccallum1998} A. McCallum and K. Nigam, "A comparison of event model for naive Bayes text classification", AAAI Workshop Learn. Text Categorization, 1998.
  \bibitem{mauri2020} A. Mauri, "alvaDesc: A tool to calculate and analyze molecular descriptors and fingerprints" in Ecotoxicological QSARs. Methods in Pharmacology and Toxicology, New York:Humana, 2020.
  \bibitem{yap2011} C. W. Yap, "Padel-descriptor: An open source software to calculate molecular descriptors and fingerprints", J. Comput. Chem., vol. 32, no. 7, pp. 1466-1474, 2011.
  \bibitem{cheng2016recommender} H. T. Cheng, L. Koc, J. Harmsen, T. Shaked, T. Chandra, H. Aradhye, G. Anderson, G. Corrado, W. Chai, M. Ispir et al., "Wide and deep learning for recommender systems", Proceedings of the 1st Workshop on Deep Learning for Recommender Systems, pp. 7-10, 2016.
  \bibitem{verhaeghe2019} H. Verhaeghe, S. Nijssen, G. Pesant, C.-G. Quimper and P. Schaus, "Learning optimal decision trees using constraint programming", CP2019, 2019.
  \bibitem{verwer2019} S. Verwer and Y. Zhang, "Learning optimal classification trees using a binary linear program formulation", 33rd AAAI Conference on Artificial Intelligence, 2019.
  \bibitem{lundberg2018consistent} Scott M. Lundberg, Gabriel G. Erion and Su-In Lee, "Consistent individualized feature attribution for tree ensembles", arXiv preprint arXiv:1802.03888, 2018.
  \bibitem{brodley1995} C. E. Brodley and P. E. Utgoff, "Multivariate decision trees", Mach. Learn., vol. 19, pp. 45-77, 1995.
  \bibitem{li2009} X.B. Li, J.R. Sweigar, J.T.C. Teng, J.M. Donohue, L.A. Thombs and S.M. Wang, "Multivariate Decision Trees Using Linear Discriminants and Tabu Search", IEEE Trans. Systems Man and Cybernetics Part A, vol. 33, no. 2, pp. 194-205, Mar. 2003.
  \bibitem{murthy1994} S. K. Murthy, S. Kasif and S. Salzberg, "A system for induction of oblique decision trees", J. Artif. Intell. Res., vol. 2, pp. 1-32, 1994.
  \bibitem{kim2001} H. Kim and W.-Y. Loh, "Classification trees with unbiased multiway splits", J. Amer. Statistical Assoc., vol. 96, pp. 589-604, 2001.
  \bibitem{loh1997} W.-Y. Loh and Y.-S. Shih, "Split selection methods for classification trees", Statist. Sinica, vol. 7, pp. 815-840, 1997.
  \bibitem{loh1998} W.-Y. Loh and N. Vanichsetakul, "Tree-structured classification via generalized discriminant analysis", J. Amer. Statist. Assoc., vol. 83, no. 403, pp. 715-728, 1988.
  \bibitem{narodytska2018} N. Narodytska, A. Ignatiev, F. Pereira and J. Marques-Silva, "Learning optimal decision trees with SAT", Proceedings of the Twenty-Seventh International Joint Conference on Artificial Intelligence IJCAI 2018, pp. 1362-1368, July 13-19, 2018, 2018.
  \bibitem{schidler2021} A. Schidler, S. Szeider. SAT-based decision tree learning for large data sets. In: Proceedings of AAAI. 2021.
  \bibitem{bessiere2009} C. Bessiere, E. Hebrard and B. O'Sullivan, "Minimising Decision Tree Size as Combinatorial Optimisation", Proc. 15th Int'l Conf. Principles and Practice of Constraint Programming, pp. 173-187, 2009.
  \bibitem{avellaneda2020} F. Avellaneda, "Efficient inference of optimal decision trees", AAAI, pp. 3195-3202, 2020.
  \bibitem{li2007} P. Li, Q. Wu and C. J. Burges, "Mcrank: Learning to rank using multiple classification and gradient boosting", Proc. Adv. Neural Inf. Process. Syst., pp. 897-904, 2007.
  \bibitem{metropolis1953equation} N. Metropolis, A. W. Rosenbluth, M. N. Rosenbluth, A. H. Teller and E. Teller, "Equation of state calculations by fast computing machines", J. Chem. Phys., vol. 21, no. 6, pp. 1087-1092, 1953.
  \bibitem{hastings1970monte} W. K. Hastings, "Monte Carlo sampling methods using Markov chains and their applications", Biometrika, vol. 57, pp. 97-109, 1970.
  \bibitem{pinheiro2006mixed} J. C. Pinheiro and D. M. Bates, "Mixed-Effects Models in S and S-Plus", New York:Springer, 2000.
  \bibitem{cobb2021scaling} A. D. Cobb and B. Jalaian, "Scaling Hamiltonian Monte Carlo inference for Bayesian neural networks with symmetric splitting" in arXiv:2010.06772, 2020.
  \bibitem{viinikka2020layering} J. Viinikka and M. Koivisto, "Layering-MCMC for Structure Learning in Bayesian Networks", Proceedings of the 36th Conference on Uncertainty in Artificial Intelligence (UAI), PMLR 124:839-848, 2020.
  \bibitem{okuyama2019binary} T. Okuyama, T. Sonobe, K.-I. Kawarabayashi and M. Yamaoka, "Binary optimization by momentum annealing", Phys. Rev. E Stat. Phys. Plasmas Fluids Relat. Interdiscip. Top., vol. 100, no. 1, Jul. 2019.
  \bibitem{stollenwerk2019quantum} T. Stollenwerk et al., "Quantum annealing applied to de-conflicting optimal trajectories for air traffic management", IEEE Trans. Intell. Transp. Syst., vol. 21, no. 1, pp. 285-297, Jan. 2020.
  \bibitem{farhi2001quantum} E. Farhi, J. Goldstone, S. Gutmann, J. Lapan, A. Lundgren and D. Preda, "A quantum adiabatic evolution algorithm applied to random instances of an NP-complete problem", Science, vol. 292, no. 5516, pp. 472-475, 2001.
  \bibitem{tao2020} M. Tao, K. Nakano, Y. Ito, R. Yasudo, M. Tatekawa, R. Katsuki, et al., "A work-time optimal parallel exhaustive search algorithm for the QUBO and the Ising model with GPU implemetation", Proc. of International Parallel and Distributed Processing Symposium Workshops, pp. 557-566, 2020.
  \bibitem{junger2021} M. Jünger. et al. Quantum annealing versus digital computing: An experimental comparison. J. Exp. Algorithmics (JEA) 26, 1–30 (2021).
  \bibitem{zaman2021pyqubo} "PyQUBO" 2022, [Online]. Available: https://pyqubo.readthedocs. io/en/latest/.  [Accessed on the 08.06.2022].
  \bibitem{ocean2022} "D-Wave Ocean Software Documentation" 2022, [Online]. Available: https://docs.ocean.dwavesys.com/en/stable/. [Accessed on the 08.06.2022].
  \bibitem{kurihara2009quantum} K. Kurihara, S. Tanaka and S. Miyashita, "Quantum annealing for clustering", In Uncertainty in Artificial Intelligence (UAI), 2009. 
  \bibitem{sasdelli2021quantum} M. Sasdelli and T. -J. Chin, "Quantum Annealing Formulation for Binary Neural Networks," 2021 Digital Image Computing: Techniques and Applications (DICTA), 2021, pp. 1-10, doi: 10.1109/DICTA52665.2021.9647321.
  \bibitem{rere2015simulated} L. M. R. Rere, M. I. Fanany and A. M. Arymurthy, "Simulated annealing algorithm for deep learning", Procedia Computer Science, vol. 72, pp. 137-144, 2015.
  \bibitem{sato2019} G. Sato, M. Konoshima, T. Ohwa, H. Tamura and J. Ohkubo, "Quadratic unconstrained binary optimization formulation for rectified-linear-unit-type functions", Phys. Rev. E Stat. Phys. Plasmas Fluids Relat. Interdiscip. Top., vol. 99, no. 4, Apr. 2019.
  \bibitem{birdal2021quantum} T. Birdal, V. Golyanik, C. Theobalt and L. Guibas, "Quantum permutation synchronization", IEEE Conference on Computer Vision and Pattern Recognition CVPR, 2021.
  \bibitem{li2020quantum} J. Li and S. Ghosh, "Quantum-soft qubo suppression for accurate object detection", In European Conference on Computer Vision (ECCV), 2020.
  \bibitem{cruz2018qubo} W. Cruz-Santos, S. E. Venegas-Andraca and M. Lanzagorta, "A QUBO formulation of the stereo matching problem for D-Wave quantum annealers", Entropy, vol. 2, 2018.
  \bibitem{noormandipour2022} M. Noormandipour and H. Wang, "Matching Point Sets with Quantum Circuit Learning," ICASSP 2022 - 2022 IEEE International Conference on Acoustics, Speech and Signal Processing (ICASSP), 2022, pp. 8607-8611, doi: 10.1109/ICASSP43922.2022.9746800.
  \bibitem{golyanik2020} V. Golyanik and C. Theobalt, "A quantum computational approach to correspondence problems on point sets", IEEE Conference on Computer Vision and Pattern Recognition CVPR, 2020.
  \bibitem{ogorman2015} B. O’Gorman, R. Babbush, A. Perdomo-Ortiz, A. Aspuru-Guzik and V. Smelyanskiy, "Bayesian network structure learning using quantum annealing", Eur. Phys. J. Special Topics, vol. 224, no. 1, pp. 163-188, 2015.
  \bibitem{shikuri2020} Y. Shikuri, "Efficient Conversion of Bayesian Network Learning into Quadratic Unconstrained Binary Optimization", arXiv preprint arXiv:2006.06926, 2020.
  \bibitem{de2011ames} D. D. Cock, "Ames Iowa: Alternative to the Boston Housing Data as an End of Semester Regression Project", Journal of Statistics Education, vol. 19, no. 3, 2011.
  \bibitem{venturclli2015} D. Venturclli, S. Mandrà, S. Knysh, B. OGorman, R. Biswas and V. Smelyanskiy, "Quantum optimization of fully connected spin glasses", Physical Review X, vol. 5, no. 3, pp. 031040, 2015.
  \bibitem{glover2018tutorial} F. Glover, G. Kochenberger and Y. Du, "A tutorial on formulating and using QUBO models", arXiv:1811.11538, Nov. 2018.




\end{thebibliography}
\end{document}